# IoMT-based Automated Leukemia Classification using CNN and Higher Order Singular Value Decomposition

Shabnam Bagheri Marzijarani
Department of Information Technology, Faculty of Mechanics, Electrical Power and Computer, Science and Research Branch, Islamic Azad University, Tehran Iran
sh1_bagheri@yahoo.com

Mohammad Zolfaghari
Department of Computer Science, University of Tehran, Kish International Campus, Iran
mmzolfaghari@ut.ac.ir

Hedieh Sajedi
Department of Mathematics, Statistics and Computer Science, College of Science, University of Tehran,
Tehran, Iran
hhsajedi@ut.ac.ir

*Abstract*—**The Internet of Things (IoT) is a concept by which objects find identity and can communicate with each other in a network. One of the applications of the IoT is in the field of medicine, which is called the Internet of Medical Things (IoMT). Acute Lymphocytic Leukemia (ALL) is a type of cancer categorized as a hematic disease. It usually begins in the bone marrow due to the overproduction of immature White Blood Cells (WBCs or leukocytes). Since it has a high rate of spread to other body organs, it is a fatal disease if not diagnosed and treated early. Therefore, for identifying cancerous (ALL) cells in medical diagnostic laboratories, blood, as well as bone marrow smears, are taken by pathologists. However, manual examinations face limitations due to human error risk and time-consuming procedures. So, to tackle the mentioned issues, methods based on Artificial Intelligence (AI), capable of identifying cancer from non-cancer tissue, seem vital. Deep Neural Networks (DNNs) are the most efficient machine learning (ML) methods. These techniques employ multiple layers to extract higher-level features from the raw input. In this paper, a Convolutional Neural Network (CNN) is applied along with a new type of classifier, Higher Order Singular Value Decomposition (HOSVD), to categorize ALL and normal (healthy) cells from microscopic blood images. We employed the model on IoMT structure to identify leukemia quickly and safely. With the help of this new leukemia classification framework, patients and clinicians can have real-time communication. The model was implemented on the Acute Lymphoblastic Leukemia Image Database (ALL-IDB2) and achieved an average accuracy of %98.88 in the test step.**

## I. Introduction

For the first time, Kevin Ashton proposed the Internet of Things (IoT) concept [1]. A system of IoT is sensors and devices connected by a cloud computing framework. For example, one of the subtypes of IoT is the Internet of Medical Things (IoMT), where medical devices are high-speed connected through a network. Clinicians and patients can establish a real-time and safe remote access connection by IoMT for gathering, processing, and transferring medical data. Many articles have been written in the IoMT and IoMT using Artificial Intelligence (AI) in recent years [2-7].

The connection between medical issues and AI methods has significantly increased with the advancement of technology. As a result, numerous medical aid systems have been proposed in the recent decade. Among these methods, diagnosing cancer is particularly important because it helps to decide efficient treatment options. Moreover, early diagnosis can be indispensable for many cancers. For example, one of the cancers that have been urgent to be identified in the beginning stage, especially in kids, is Acute Lymphocytic Leukemia (ALL). It is a kind of hematic disease resulting from an abnormality in the production of immature White Blood cells (WBCs) in the bone marrow [8]. Because many different types of leukemia do not show obvious symptoms at the onset of the disease, leukemia may be diagnosed accidentally during a physical examination, including blood tests and bone marrow biopsies. Because humans are directly involved in it, mistakes may occur during diagnosis. Blood microscopic images cells analysis is the most economical way to perform primary screening for patients with leukemia. An inexpensive automatic, robust, and real-time system is required to prevent the operator from being affected.

Many automatic methods based on AI have been introduced to diagnose diseases. For example, Computer-aided Diagnostic (CAD) systems have been developed using image processing and computational intelligence techniques. Recently, Convolutional Neural Networks (CNN) have received more attention for image classification tasks than others [9-12]. Deep learning (DL) and machine learning (ML) use multi-layered artificial neural networks to provide precision and progress in image classification, object recognition, speech recognition, language translation, and more. From a learning point of view, DL differs from traditional ML techniques. They can learn from given data without introducing manipulated rules or human knowledge. Their highly flexible architectures can learn directly from raw data, and if more data is prepared, their predictive accuracy can be increased. Different structures are provided for DL depending on the various applications. CNN is one of the most widely used DL network structures [13].

CNNs, regularized version of a multilayer perceptron, are inspired by biological processes whose connection pattern between neurons is similar to that of the animal visual cortex. As a result, CNNs have led to considerable success in implementing real-world applications. Such applications include image classification, image detection, pattern recognition, vocal recognition, Natural Language Processing (NLP), video analysis, etc. [14].

CNNs use a hierarchical pattern in the data and collect more complex patterns using smaller, simpler ones. Detailed image processing shows that convolution is an impressive way to extract image features by reducing data dimension and redundant data, also called a feature map. Each kernel acts as an attribute identifier, filtering where that attribute is in the original image. Finally, it produces a map whose altitude shows how these features are distributed. Image data has been transmitted forward to the convolution and pooling layers. Therefore, each layer of CNN includes feature extraction from the previous steps. As a result, it contains information that can be used as input to any classifiers. Typically, one of the workhorse layers in CNN called Fully Connected (FC) is of interest to researchers as an image feature extractor [15-18]. We focus on the classification step in CNN in this study. The last layer of CNNs can be considered a linear classifier. But, it is not the optimal classifier. Therefore, optimal classifiers, such as Support Vector Machine (SVM) and CNN [1], are the most common methods.

In this research, we have applied a robust classifier, Higher-Order Singular Value Decomposition (HOSVD), based on multi-linear algebra concepts. Compared to classic and DL classifiers, the HOSVD classifier has faster training speeds, fewer tuning parameters, shorter running time, and higher training accuracy. Also, the method is developed into a framework base on the IoMT.

This paper is organized as follows: Related works are described in Section II. The suggested method is introduced in Section III. Used database and experimental results have been shown in Section IV, and conclusions and perspectives on future work are provided in Section V.

## II. RELATED WORKS

Several methods have been proposed for diagnosing leukemia over the years, and some of these methods offer solutions for classifying two common types of leukemia: ALL and healthy. In this section, previous related works for categorizing ALL and healthy cells have been studied and reviewed. The research results are summarized in Table I.

TABLE I. RELATED WORKS

| **Ref** | **Year** | **method** | **Feature(s)** | **Classifier(s)** |
|---|---|---|---|---|
| [19] | 2022 | Deep | CNN (SENet) | CNN |
| [20] | 2021 | Deep | CNN (AlexNet) | SVM |
| [21] | 2020 | Deep | Color histogram, LDP | GrayJOA and CNN |
| [22] | 2019 | Deep | CNN | CNN |
| [23] | 2019 | Deep | Color histogram, LDP | CNN |
| [24] | 2018 | Classic | Shape, color, and texture | KNN, SVM, NB, DT |
| [25] | 2018 | Deep | CNN | CNN |
| [26] | 2018 | Classic | Texture, statistical, and geometrical | C-KNN |
| [27] | 2014 | Classic | LDP | SVM |

According to Table I, many previous related works have used CNN for feature extraction and classification. This article is introduced a new CNN architecture for binary leukemia classification. First, the features are extracted with a CNN, and then multi-linear algebraic concepts perform classification. In the following, we describe the proposed method in the next section.

## III. PROPOSED METHOD

How to find effective features is the core issue in the diagnosis of leukemia. DL is a newly developed approach for pattern recognition inspired by the human visual cortex. Involving a class of models, DL tries to learn multiple levels of data representation. Extraction of abstract and invariant features results from this practical learning for various tasks such as classification and identification. This paper is used a specific type of DL called CNN as a feature extractor alongside tensor decomposition as a classifier. The detail of both of them will be described in the following.

### *A. CNN architecture*

CNN is becoming a hotspot in ML and AI studies. Receptive fields (windows) and kernels (filters or cores) are the foundation of the convolutional and pooling layer of the CNN architecture. These technologies enable CNN to extract specific features from different input data space locations, thus tolerating shift, scale, and distortion invariance. Such advantage of CNN in exploring representative features from various image data parts can be taken to the leukemia classification. Given each pixel input as an image directly embedded in a deep CNN-centric image classification, a special network designed for leukemia image classification tasks can extract high-level features from raw data. This scientific study shows the properties obtained in the near-end convolution.

### *B. Higher-Order Singular Value Decomposition (HSOVD)*

Higher-Order Singular Value Decomposition (HOSVD) was first introduced by Hitchcock [28] and then developed [29-32]. It is a multi-linear algebra method related to the tensor concept, a specific orthogonal tucker decomposition, and practical in different applications such as classification problems [33, 34].

Assume we have a two-dimension matrix, and it is shown in equation (1):

$$F \in R^{m \times n}, \ m \geq n \qquad (1)$$

Equation (2) is displayed the singular value decomposition (SVD) of this matrix:

$$F=U\Sigma V^{T}\ ,\ U\in R^{m\times m},\ \Sigma\in R^{m\times n},\ V\in R^{n\times n} \qquad (2)$$

where U and V are orthogonal matrices, $\Sigma$ is a diagonal matrix. Therefore, the formula for the second-order tensor is shown in equation (3):

$$F=\Sigma X_1 U X_2 V \qquad (3)$$

Now, the Higher-order SVD (HOSVD) of a tensor can be written as the following equation:

$$T=T' X_1 U X_2 V X_3 W\ ,\ U\in R^{I\times I},\ V\in R^{J\times J},\ W\in R^{K\times K} \qquad (4)$$

where U, V, and W are orthogonal matrices. $T'$ is represented as a real tensor with the same T dimension and is satisfied.

The steps of the HOSVD classifier are as bellow:

- The tensors from the last convolution layer are sorted into the tensors with the same kinds.
- The HOSVD of the tensors is calculated.
- The basis normalized matrices are computed.

### *C. The proposed network and IoMT framework*

We designed a specific CNN for automatic leukemia classification. Fig. 1 shows the architecture of the proposed network.

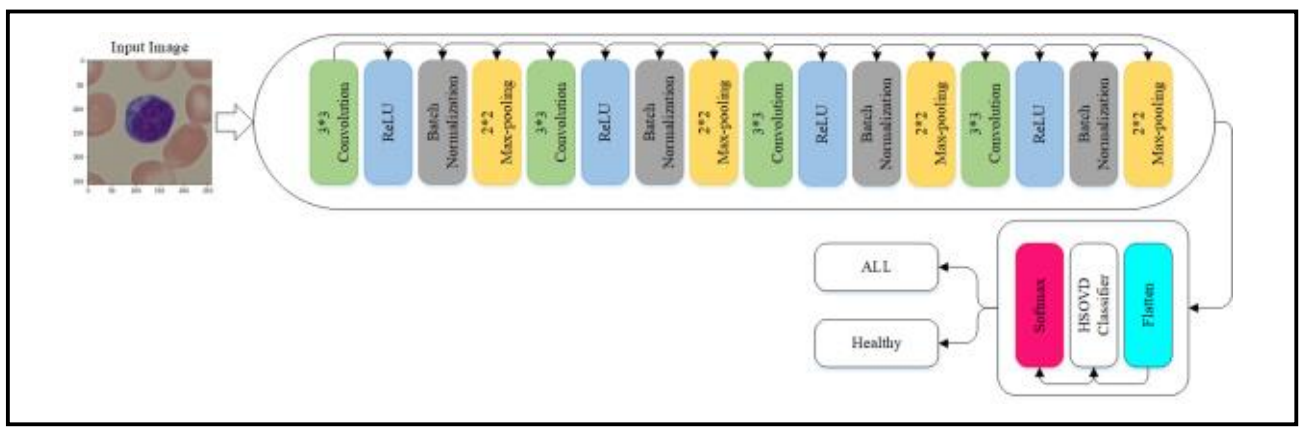

Fig. 1. The architecture of the proposed network

The network can develop into the IoMT environment for real-time connection between hematologists and patients. The framework of the IoMT-based automated leukemia classification is displayed in Fig. 2.

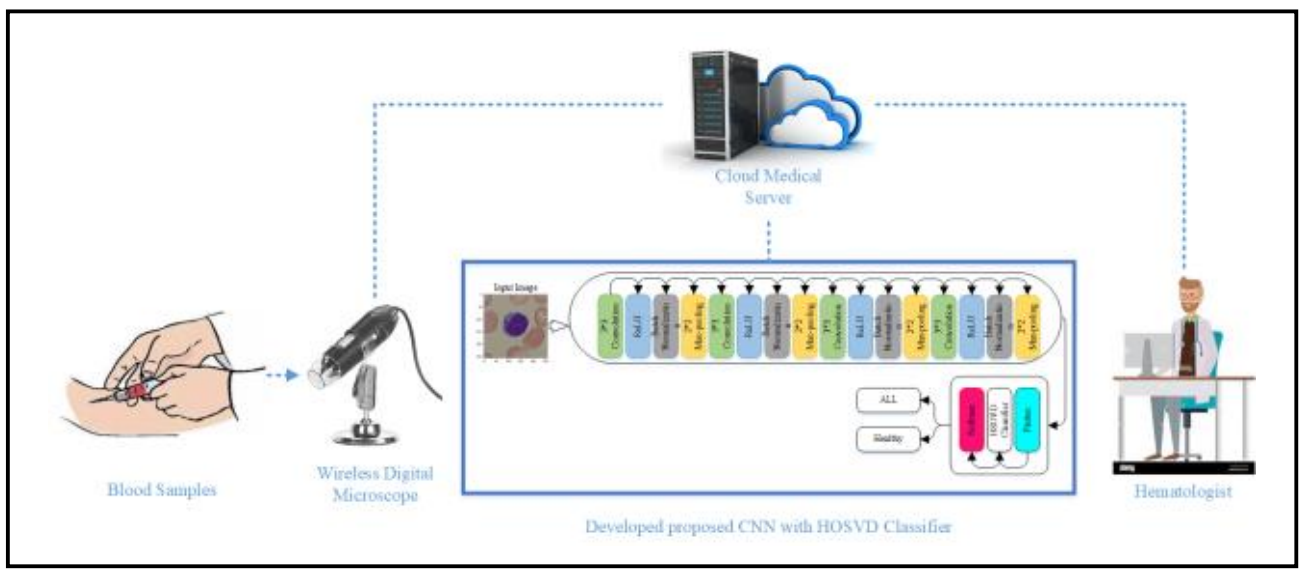

Fig. 2. The framework of The IoMT-based automated leukemia classification

First, the blood samples of persons are taken and collected. Then, the collected samples are transferred to a cloud medical server with a wireless digital microscope. After that, the extracted information from the previous step are sent to the proposed network for automatic classification. Finally, the network results were sent back to the cloud medical server and then forwarded to the monitor of the computer or laptop of hematologists. As explained, all stages of identifying people with leukemia and healthy ones are done automatically and with the help of IoMT. This classification has advantages such as high speed and accuracy and saving memory consumption, performed with minimal human intervention.

## IV. EMPLOYED DATABASE AND EXPERIMENTAL RESULTS

The database used in this study is ALL-IDB2, proposed by Labati et al. [35], designed for the classification task. Because of being standard and free available, It is an appropriate database for automatically classifying ALL and healthy cells. The ALL-IDB2 comprises 260 microscopic blood images, including 130 ALL and 130 healthy cells [36]. The original size of each image is 257 × 257 pixels. Fig. 3 shows samples of ALL and healthy cells of the database.

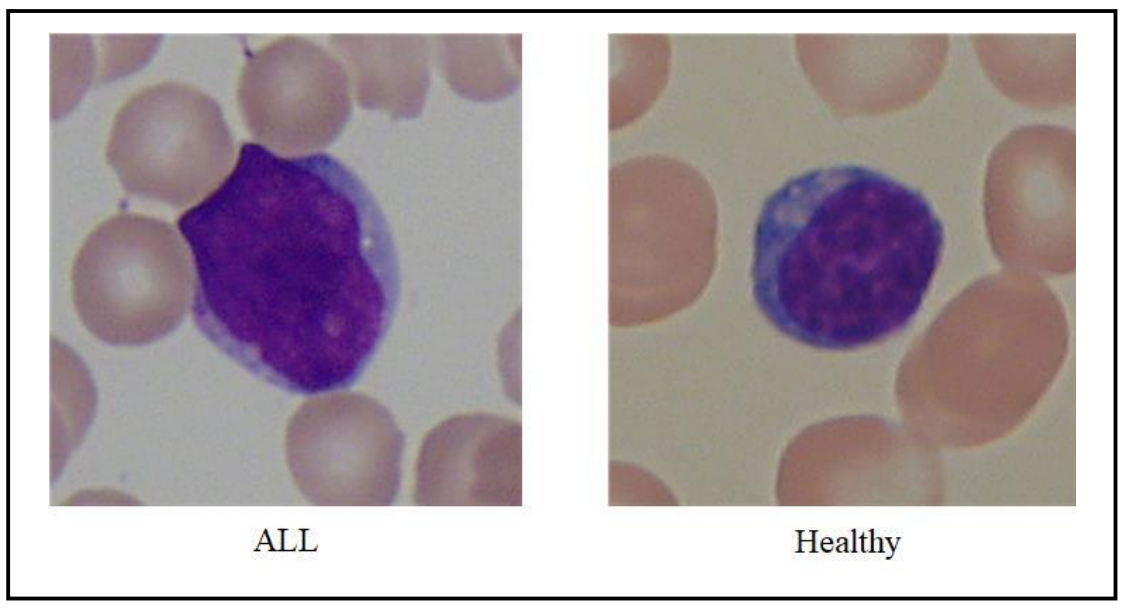

Fig. 3. Sample of ALL and healthy cells of the ALL-IDB2 database

The model was implemented on 80% of database samples with a learning rate of 0.001 for 20 epochs. It achieved an average accuracy of 98.88% and a standard deviation of 0.009 in the test step. To compare the performance of the HOSVD with other state-of-the-art classifiers, we replace CNN, Extreme Learning Machine (ELM) [37], K-Nearest Neighbor (KNN) [38], and SVM with it. Table II. shows the average accuracy and standard deviation of classifiers. As can be depicted from Table II, not only the HOSVD-based classifier has achieved the best average accuracy, but also it has attained the lowest standard deviation among other classifiers.

TABLE II. RESULTS OF DIFFERENT CLASSIFIERS (%)

| **Classifier/metric** | CNN | SVM | ELM | KNN | HOSVD |
|---|---|---|---|---|---|
| Average accuracy | 97.75 | 98.12 | 98 | 97.75 | **98.88** |
| Standard deviation | 0.015 | 0.0174 | 0.0154 | 0.0175 | **0.009** |

To analyze the differences between the classifiers, we also display the graphical examination of the Analysis of Variance (ANOVA) test for classifiers results in Fig. 4. ANOVA is introduced by Ronald Fisher [39]. This concept is a statistical equation employed to compare variances across the average of different groups. Maitra and Mathew presented ANOVA in statistical image processing [40]. As we can see in Fig. 4, with the help of ANOVA, the HOSVD classifier has the best average rather than the others.

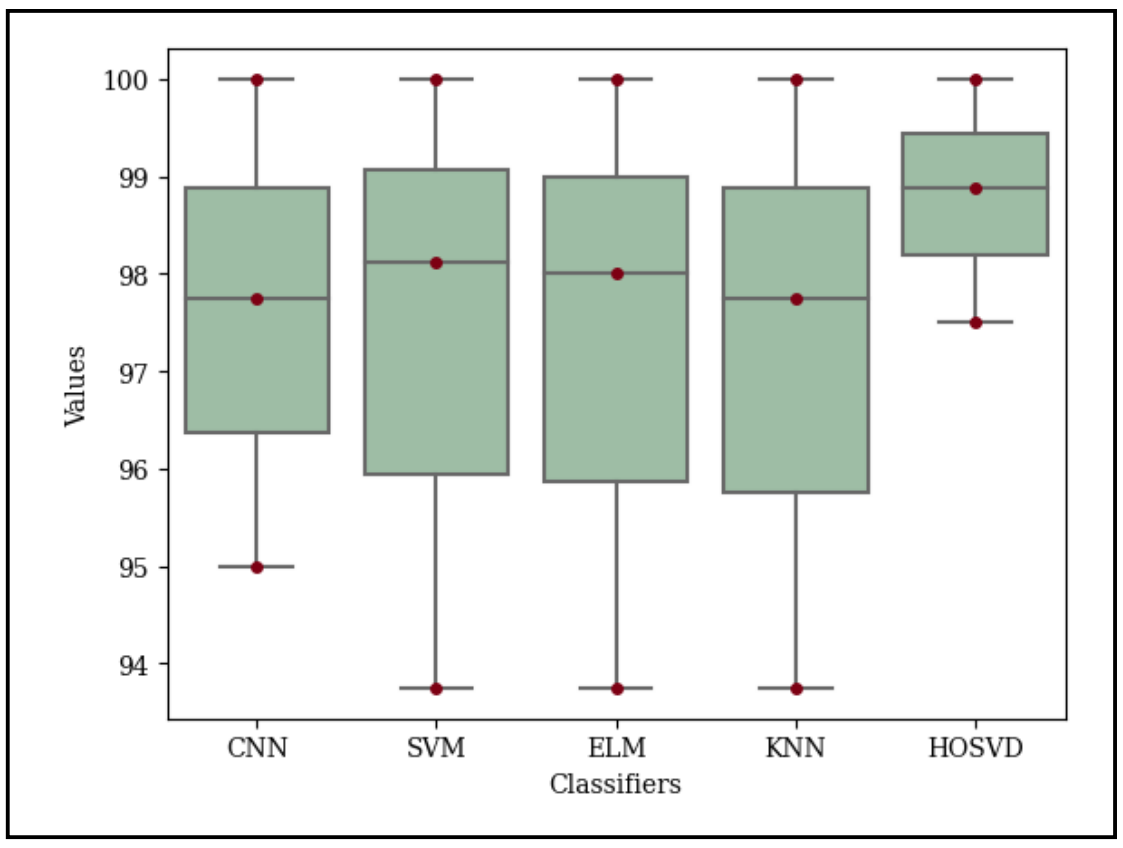

Fig. 4. Result of ANOVA test

The suggested method is compared with the other state-of-the-art methods in which standard ML classifiers are used for classification. Table III shows that the proposed CNN-HOSVD offers the highest accuracy of 98.88% and goes beyond the base approaches. The classification accuracy rate proves the proposed method is promising for diagnosing leukemia.

TABLE III. COMPARISON BETWEEN THE SUGGESTED MODEL WITH OTHER STATE-OF-THE-ART RELATED METHODS

| **Ref** | **Year** | **Used Database** | **Classifier(s)** | **Accuracy (%)** |
|---|---|---|---|---|
| [19] | 2022 | ALL-IDB2 | CNN | 98.16 |
| [20] | 2021 | ALL-IDB2 | SVM | 96.15 |
| [21] | 2020 | ALL-IDB2 | GrayJOA and CNN | 93.5 |
| [22] | 2019 | ALL-IDB2 | CNN | 88.25 |
| [23] | 2019 | ALL-IDB2 | CNN | 98.7 |
| [24] | 2018 | ALL-IDB2 | KNN, SVM, NB, DT | 96.42 |
| [25] | 2018 | ALL-IDB2 | CNN | 96.6 |
| [26] | 2018 | ALL-IDB2 | C-KNN | 96.25 |
| [27] | 2014 | ALL-IDB2 | SVM | 89.72 |
| Our model | 2023 | ALL-IDB2 | HOSVD | **98.88** |

## V. Conclusion and Future Work

This study presents a new approach to leukemia diagnosis rather than linking state-of-the-art methods. The suggested model extracts the features using CNN, and the HOSVD, a multi-linear algebra concept, performs classification. In addition, we introduced a framework based on IoMT for leukemia classification that cause peoples and hematologists to relate together in real time. Comparing the category of the proposed method with other experiences shows that the new approach to diagnosing leukemia has reached the highest average accuracy among other proposed methods and has a minimum standard deviation. Therefore, it can be considered a robustness method for further pattern recognition.

For future work, the suggested network can implement in similar databases and for multi-class classification. Furthermore, it will act as an initiative to combine CNN with other subspace learning algorithms. Also, we recommend that the IoMT-based framework introduced in this study be used experimentally in hospitals and medical centers.